\RequirePackage{fix-cm}
\documentclass[smallextended]{svjour3}       
\smartqed  
\usepackage{graphicx}
\usepackage[flushleft]{threeparttable}
\usepackage{amssymb}
\usepackage{amsmath}
\usepackage{multirow}
\usepackage{subfigure}
\usepackage{gensymb}
\usepackage{lscape}
\usepackage{booktabs}
\usepackage{tabularx}      
\begin{document}\sloppy

\title{Generating protected fingerprint template utilizing coprime mapping transformation
	\thanks{The authors are thankful to SERB (ECR/2017/000027), Deptt. of science \& Technology, Govt. of India for providing financial support to carry out this research work. Also, we would like to thank Indian Institute of Technology Indore for providing laboratory facilities.}
}

\titlerunning{Generating protected fingerprint template utilizing coprime mapping transformation}        

\author{Rudresh Dwivedi \and Somnath Dey}


\institute{R. Dwivedi \at
              Discipline of Computer Science and Engineering,\\ Indian Institute of Technology Indore, Indore, 453446, India\\
              Tel.: +91-971-3888726\\
              \email{phd1301201006@iiti.ac.in}           
           \and
           S. Dey \at
              Discipline of Computer Science and Engineering,\\ Indian Institute of Technology Indore, Indore, 453446, India\\
}

\date{Received: date / Accepted: date}

\maketitle

\begin{abstract}
The identity of a user is permanently lost if biometric data gets compromised since the biometric information is irreplaceable and irrevocable. To revoke and reissue a new template in place of the compromised biometric template, the idea of cancelable biometrics has been introduced. The concept behind cancelable biometric is to irreversibly transform the original biometric template and perform the comparison in the protected domain. In this paper, a coprime transformation scheme has been proposed to derive a protected fingerprint template. The method divides the fingerprint region into a number of sectors with respect to each minutiae point and identifies the nearest-neighbor minutiae in each sector. Then, ridge features for all neighboring minutiae points are computed and mapped onto co-prime positions of a random matrix to generate the cancelable template. The proposed approach achieves an EER of 1.82, 1.39, 4.02 and 5.77 on DB1, DB2, DB3 and DB4 datasets of the FVC2002 and an EER of 8.70, 7.95, 5.23 and 4.87 on DB1, DB2, DB3 and DB4 datasets of FVC2004 databases, respectively. Experimental evaluations indicate that the method outperforms in comparison to the current state-of-the-art. Moreover, it has been confirmed from the security analysis that the proposed method fulfills the desired characteristics of diversity, revocability, and non-invertibility with a minor performance degradation caused by the transformation.
\keywords{Biometric \and Cancelable biometrics \and Fingerprint verification \and Template protection}

\end{abstract}

\section{Introduction}
\subsection{Background}
Compromise of original biometric information causes permanent identity theft as it is intrinsically linked to the user. In thee literature, there are various security concerns associated with the usage of biometric information across different applications \cite{ratha,anneal}. Ratha et al. \cite{ratha} first proposed three cancelable transformations namely cartesian, polar and functional transformations. Shin et al. \cite{etri} applied dictionary attack onto these transformations to derive a pre-image and found that the transformation is vulnerable to attacks. Therefore, biometric template protection is the utmost need in recent years. There are two well-known mechanisms to provide template protection i.e. Biometric cryptosystem and Cancelable biometrics. The biometric cryptosystem reforms the original template and generates poor matching rate. Hence, many researchers utilized cancelable biometric-based transformations to provide template protection. It states that a transformed template is required to be stored instead of the original biometric template. The transformation depends upon an irreversible or non-invertible function such that it is infeasible to invent the original biometric template even if the attacker gets access to the protected template and transformation key. In case of compromise, a new template can be derived by modifying the transformation key used for protected template generation. The cancelable transformation must follow the four desired characteristics:
\begin{enumerate}
	\item Non-invertibility: It should be computationally hard enough to construct the original template from the transformed template. This prevents the recovery of original biometric information by an imposter. 
	\item Diversity: Identical cancelable template should not be used in the different applications to avoid cross-matching of the stored template. 
	\item Revocability: The transformation should be able to derive numerous protected templates from the same biometric input and there should be immediate revocation in case of compromise.
	\item Performance: The transformation should not exhibit significant performance degradation.
\end{enumerate}

\subsection{Existing approaches}
In literature, authors have proposed several methods for cancelable template generation in recent years. Ratha et al. \cite{ratha} have introduced cartesian, polar, and functional transformation for fingerprint template protection. Das et al. \cite{das} evaluated nearest neighbor distance from minutiae points to the core point. Next, they constructed a graph structure which can be revoked using a PIN. However, the approaches introduced by Ratha \cite{ratha} and Das \cite{das} et al. require pre-alignment of different fingerprints based on singular/core point and detection of the core point is not always feasible.

Lee et al. \cite{lee} introduced a technique which pre-aligns minutiae points and maps into a 3-D array based on the minutia's orientation and positional difference. Next, the array is visited in sequential order to generate a bit string. The derived bit string is exploited to random permutation based on a user-specific key and minutiae type. An alignment-free protected template design method is proposed by Wang et al. \cite{ditom} where pair-minutiae vectors are quantized, indexed and converted to bit-string. Then, discrete Fourier transform (DFT) is applied to generate a complex vector which is further fused with a user-specific key. In their another work, Wang et al. \cite{blind} proposed a method to protect the bit-string derived using the method proposed in \cite{ditom}. The bit-string is utilized as an input to FIR filter with a user-specific key. Moujahdi et al. \cite{shell} presented a unique transformation i.e. fingerprint shell where they computed the distance from the singular point to all other minutiae. Next, a user-specific key is summed up to the computed distances and are sorted in ascending order to generate a spiral curve structure. Jin et al. \cite{hamem} introduced a method where minutiae triplet features are computed, i.e. three sides, three internal angle and relative orientation. Further, a random sequence is projected onto features to derive the protected template. Jin et al. \cite{jintee} proposed a protected fingerprint template generation scheme where minutiae information is first aligned and shifted to a new position. Next, the polar transform is utilized, and bit-string is derived by applying quantization. Further, the random permutation is exploited to generate the protected template. The approaches proposed in \cite{shell,lee,ditom,blind,jintee,hamem} results to performance degradation if the user-specific key gets compromised. Further, Wang et al. \cite{hadamard} presented a method where a bit string is derived by utilizing partial Hadamard transform. In BioHashing based approaches \cite{biohash,hash,hash2}, a protected template is derived after discretizing the inner product of the biometric features with the projection matrix. BioHashing and its variants \cite{biohash,hash,hash2}, are proved to be impractical if the unique seed is compromised. Sandhya et al. \cite{sandhya} proposed two different algorithms to design cancelable fingerprint template where Delaunay triangle based features are incorporated. However, the performance gets degraded in case of poor quality images. Abe et al. \cite{abe} proposed a method on template generation where bloom filter is applied onto bit-string derived by minutiae relation code. The scheme is susceptible to inversion attack if parameters of bloom filter get compromised. 

Cappelli et al. \cite{mcc} proposed a state-of-the-art minutiae representation MCC (Minutiae cylinder Code) where a 3-D cylindrical structure is framed in the vicinity of each minutiae neighborhood. Thereafter, Ferrara et al. \cite{pmcc} introduced a method namely protected-MCC (P-MCC) where they applied binary-KL projection for each MCC templates to alleviate privacy issues over non-invertibility in MCC \cite{mcc}. However, it has been further investigated that the P-MCC approach is irrevocable to some extent. To provide revocability, Ferrara et al. \cite{2pmcc} presented two-factor protected Minutiae Cylinder-Code (2P-MCC) where partial permutation is performed using a secret key over the cylinders in P-MCC. Recently, Rathgeb et al. \cite{rathir} proposed a theoretical estimation of the irreversibility of the generic protected biometric templates. They evaluated the complexity of inverting the protected template by quantifying the security provided by the different approaches. Thereafter, a novel general framework for the evaluation of unlinkability among biometric templates is proposed by Gomez-Barrero et al. \cite{unlink}. Also, a protocol is defined to analyze the correlation between different templates of same and different subjects.

\subsection{Contributions}
To mitigate the different concerns raised by afore-stated approaches, we have proposed a novel protected template generation scheme for fingerprint biometric which is based on coprime mapping transformation. We highlight the contributions of this work as follows:

\begin{enumerate}
	\item In this work, coprime mapping transformation has been applied over ridge features which deal with rotation, scale and translation distortions in the input fingerprint image effectively.
	\item The applied transformation does not depend on prior alignment based on the singular/core points as it is not always possible to extract the singularities in poor quality images.
	\item The nearest neighbor transformation is applied in the vicinity of each minutia to compute a fixed length descriptor instead of fixed-radius transformation. This would prevent performance degradation caused due to the border minutiae points.
	\item We have evaluated our method against the desirable characteristics of template security schemes, i.e. non-invertibility, revocability, and diversity.
	\item The recognition performance of the proposed approach is tested on two different benchmark databases i.e. FVC2002 and FVC2004. The experimental evaluations indicate that our method outperforms in comparison to the current state-of-the-art.
\end{enumerate}

This article is an extension of our earlier work \cite{mymike}. In the previous work, we proposed a technique for generation of cancelable fingerprint template. However, the previous work does not include a rigorous experimental analysis concerning accuracy and attack analysis. Furthermore, the earlier method was tested with only FVC2002 database \cite{fvc}. In this work, experiments have been performed onto all four datasets i.e.DB1, DB2, DB3 and DB4 of FVC2004 database. Further, experimental results are also compared with existing methods to determine the robustness of the proposed method. Additionally, we have performed a rigorous security analysis against different possible attacks such as brute-force, pre-image and annealing attack in this work. Finally, we have enhanced our literature review by adding few relevant existing approaches describing current advancement in the area. 

The remainder of the paper is organized as follows. Section 2 presents the proposed protected template generation method in detail. Experiments and comparisons are demonstrated in Section 3. Section 4 analyzes the security and privacy of the transformation. Section 5 concludes the paper.

\section{Proposed scheme}

The proposed method comprises of three major steps which include the preprocessing and minutiae extraction, feature extraction, and cancelable template generation. The overall workflow of the proposed method is depicted in Fig. 1.  

\begin{figure}[t]
	\begin{center}
		\includegraphics[width=\linewidth,height=6.5cm]{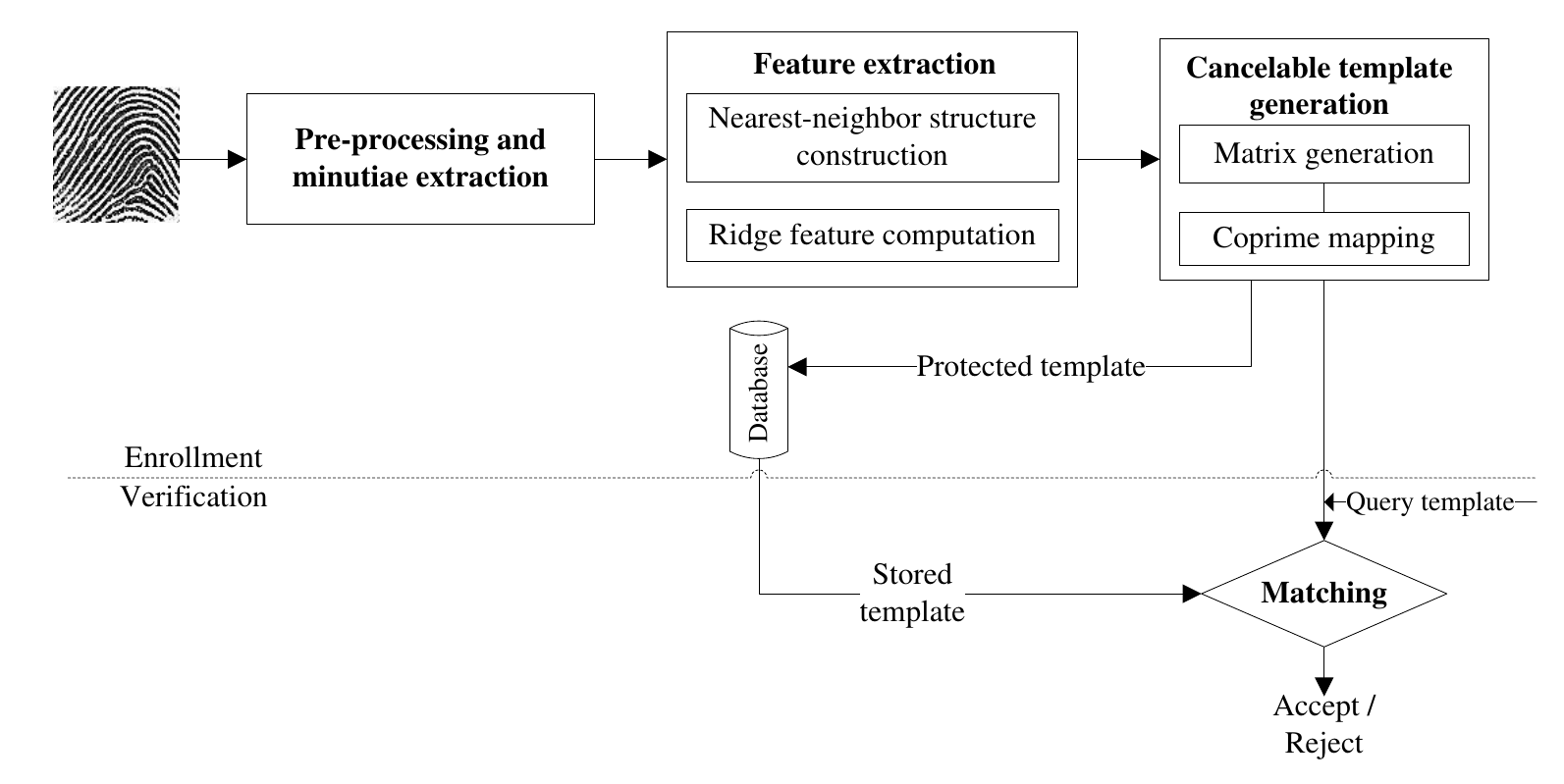}
	\end{center}
		\caption{Block diagram of the proposed method}
\end{figure}

\subsection{Pre-processing and minutiae extraction}

In our method, pre-processing and minutiae point extraction is carried out using the scheme described in \cite{joshua}. Here, the minutiae points are denoted as:
\begin{align}
V_{up}&=\left \{ m_{i} \right \}_{i=1}^{n}  \nonumber  \\        
m_{i}&=\left ( x_{i}, y_{i}, \theta_{i} \right )  
\end{align}
where, $V_{up}$ represents the set of raw minutiae points extracted from an input fingerprint, $i^{th}$ minutiae point is denoted by $m_{i}$ out of total \textit{n} minutiae points. $(x_{i},y_{i})$ and $\theta_{i}$ denotes  the coordinate position and orientation for the $i^{th}$ minutiae point. We also obtain a thinned fingerprint image which is further used for invaqriant feature extraction.

\subsection{Feature extraction}
Invariant feature computation from a fingerprint image is an utmost need since performance may degrade due to rotation, translation and scale uncertainties caused at the time of acquisition. In this work, we calculate ridge features to deal with these deformations present in the input fingerprint image. In this work, feature extraction is carried out using two steps: nearest-neighbor structure construction and ridge feature computation. 

\subsubsection{Nearest-neighbor structure construction:}
Following the preprocessing and minutiae extraction, we achieve a thinned output image and minutiae information. Afterward, we consider one of the minutiae from the minutiae set \textit{$V_{up}$} as a reference minutia. Next, a nearest-neighbor structure is formed around the reference minutia based on the ridge coordinate system as shown in Fig. 2(a). The ridge coordinate system assigns the reference axis such that it coincides with the orientation of the reference minutiae. Further, the fingerprint region is divided into `\textit{s}' sectors of equal angular width utilizing ridge coordinate system as illustrated in Fig. 2(a).
\begin{figure}[ht]

	\begin{center}
		\subfigure[Nearest-neighbor structure for $s$ = 8]{
			\includegraphics[height=6cm,width=0.40\textwidth]{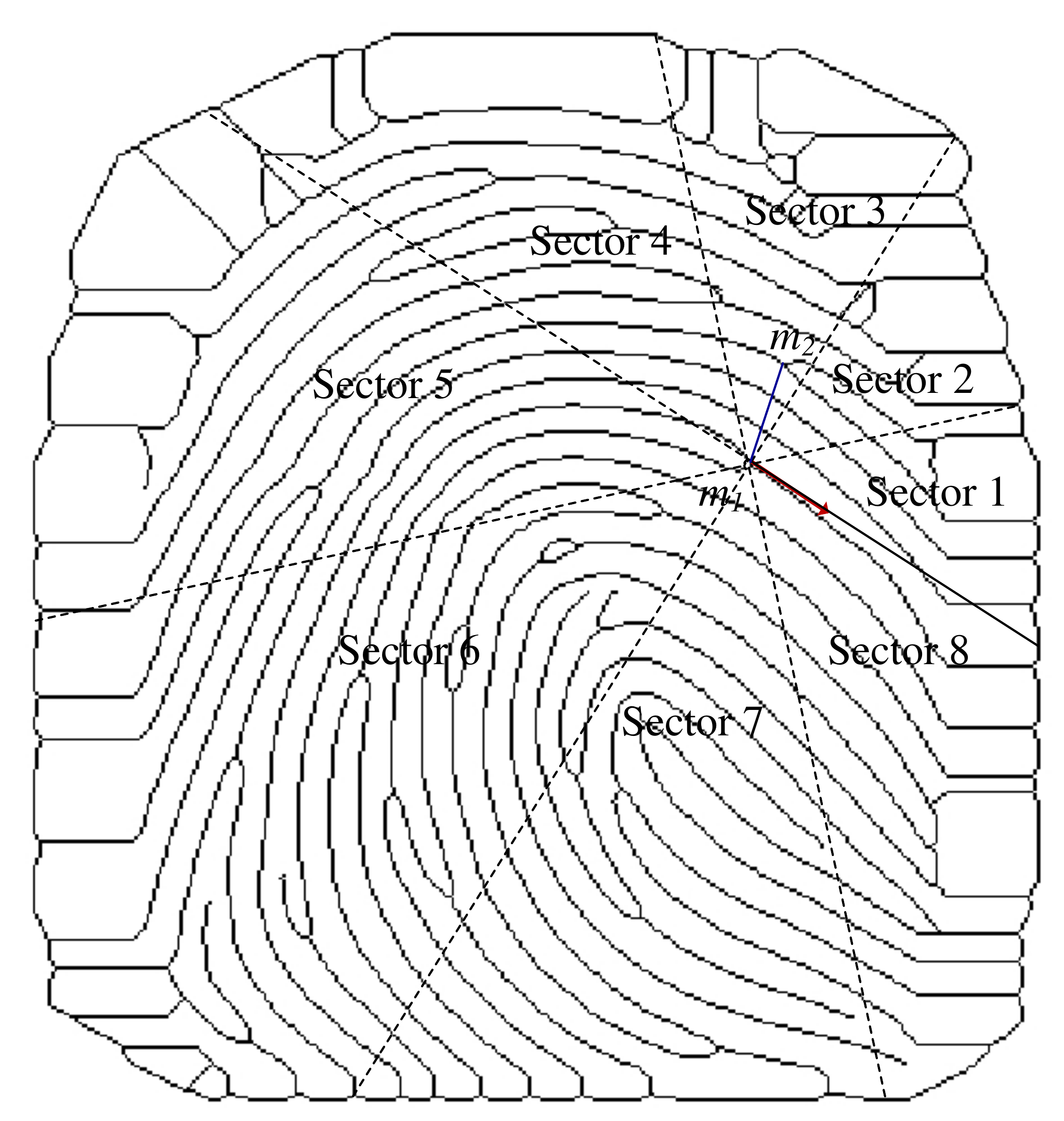}
		}
		\subfigure[Ridge-based feature computation]{
			\includegraphics[height=6cm,width=0.45\textwidth]{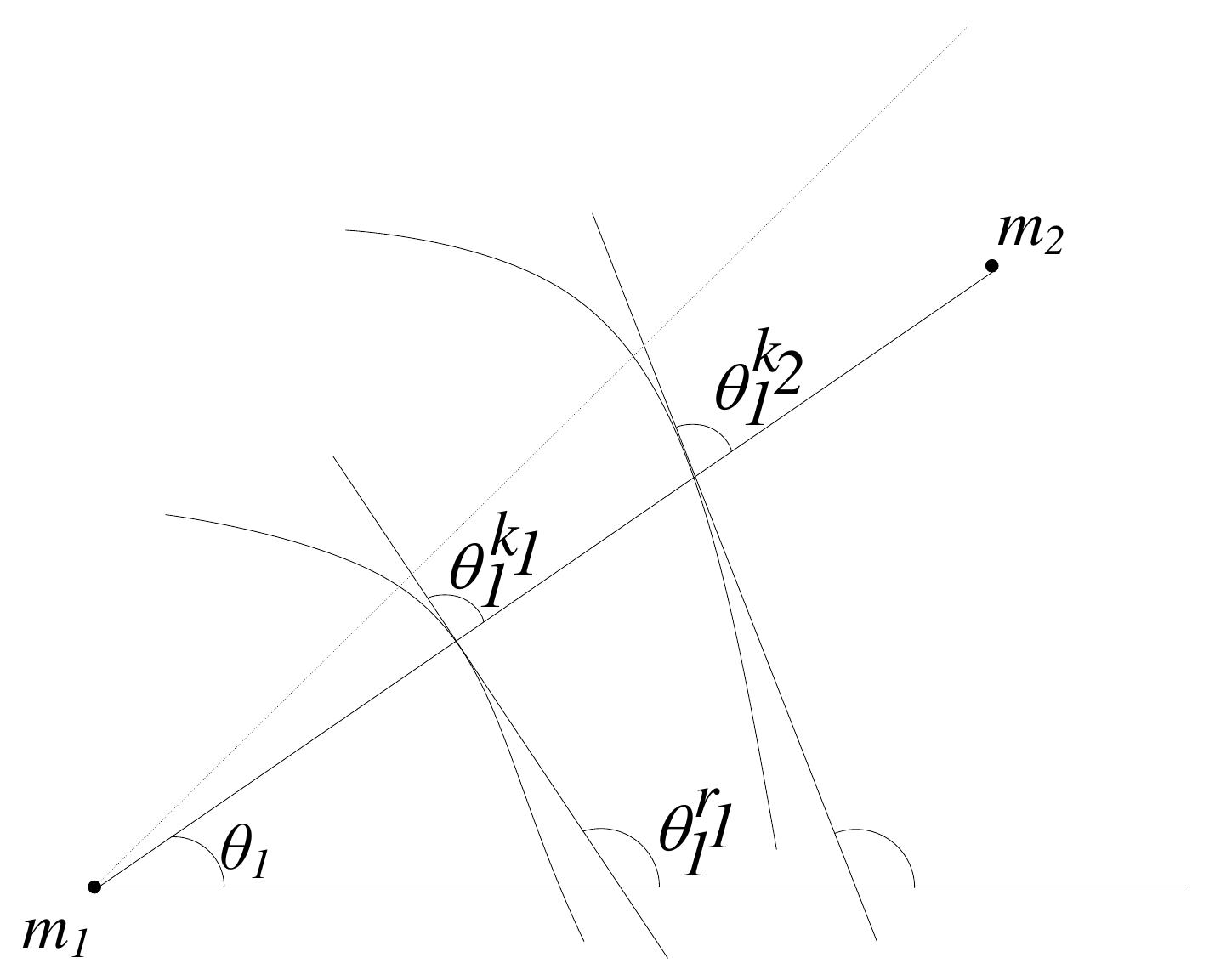}
		}
	\end{center}
	\caption{Feature extraction}
\end{figure}

\subsubsection{Ridge feature computation:}
We consider ridge count and average ridge orientation as an invariant feature in this work. The ridge features are calculated between reference minutiae and nearest minutiae in each sector considering each minutia as the reference. Ridge count is evaluated by counting the ridges between reference minutiae and nearest neighbor minutia. For example, ridge count between two minutiae points ( say $m_{1}$ and $m_{2}$ ) is 2 as shown in Fig. 2(b). To compute ridge orientation, the angle subtended by the tangent line and the straight line connecting two minutiae points is measured for each ridge crossing. For example, the orientation of the first ridge in the first sector as shown in Fig. 2(b), $\theta _{1}^{k1}$ can be evaluated as:

\begin{align*}
\textrm{for sector 1:}  \ \ 
\theta _{1}^{k1}= \theta _{1}^{r1}- \theta_{1}
\end{align*}
where $\theta_{1}$ is the slope of the straight line starting from reference minutiae to neighboring minutia in the first sector. $\theta _{1}^{r1}$, is the angle spanned by a tangent line from the intersection point of first ridge and reference axis. Similarly, we compute the orientation of the second ridge, $\theta _{1}^{k2}$ and evaluate the mean ridge orientation for example shown in Fig. 2(b). The mean ridge orientation for each sector can be expressed as defined in Eq. (2).
\begin{equation}
\textrm{for $i^{th}$ sector:}\ r_{or_{i}}
= \frac{\left ( \theta _{i}^{r1}- \theta_{i} \right )+ \left ( \theta _{i}^{r2}- \theta_{i} \right )  + ........+ \left ( \theta _{i}^{rNr_{i}}- \theta_{i} \right )}{Nr_{i}} 
\end{equation}

where $Nr_{i}$ denotes the total number of ridges between the nearest minutiae and the reference minutiae in the $i^{th}$ sector. The ridge features are stored in a 2-D matrix ($F$). For example, if a fingerprint contains $n$ number of minutiae points, the feature matrix $F$ will contain $n\times2s$ entries containing $s$ ridge count and $s$ average ridge orientation if the fingerprint image is divided into $s$ number of sectors. A value zero is assigned to the ridge features if no minutia point is located in that sector. At the time of matching, we do not consider the sectors with no minutiae point.

\subsection{Cancelable template generation}
The generation of cancelable fingerprint template involves two steps: matrix generation and co-prime mapping. 
\subsubsection{Matrix generation:}
The feature matrix is mapped into a high-dimensional matrix to generate the protected template. For this purpose, a random matrix $CanTemp$ of size $T\times T$ is generated using a seed ($\rho$). The value of $T$ is equal to $n\times2s$ where $n$ and $s$ are the total number of minutiae points in the input fingerprint image and the number of sectors around a reference minutiae, respectively.

\subsubsection{Co-prime mapping:}
We map the feature matrix $F_{n\times2s}$ into $CanTemp$  such that there will be no overlapping. To perform this, we map the ridge features $F$ onto coprime positions at $T$ places of $CanTemp$. Rest of the entries of the matrix are filled with some random data. The following four keys are utilized for mapping:

\begin{enumerate}
	\item $k_{1}$ : initial row position \ \	\item $k_{2}$ : initial column position  \ \
	\item  $k_{3}$ : number of row jump from initial position	\item $k_{4}$ : number of column jump from initial position 
\end{enumerate}

The start position is selected on the basis of the user-specific key. We start at position $(k_{1}, k_{2})$ in matrix $CanTemp$. The next position ($NP$) is evaluated based on the row and column jump from the initial position using the following relation described in Eq. (3) and (4):

\begin{center}
	\begin{equation}
	NP_{i}=\left\{\begin{matrix}
	k_{1}+k_{3} & if \left ( k_{1}+k_{3}\leqslant T \right )\\ 
	k_{1}+k_{3}-T & if \left ( k_{1}+k_{3} > T \right )
	\end{matrix}\right.
	\end{equation}
\end{center}

\begin{center}
	\begin{equation}
	NP_{j}=\left\{\begin{matrix}
	k_{2}+k_{4} & if \left ( k_{2}+k_{4}\leqslant T \right )\\ 
	k_{2}+k_{4}-T & if \left ( k_{2}+k_{4} > T \right )
	\end{matrix}\right.
	\end{equation}
\end{center}

The coprime mapping avoids overlapping in the matrix. Further, we select the value of $k_{3}$ and $k_{4}$ such that both should be co-prime with $T$ as defined in Eq. (5).

\begin{center}
	\begin{equation}
	\begin{matrix}
	GCD\left ( k_{3}, T \right )=1 & \forall  & k_{3}\epsilon [2,T] \\ 
	GCD\left ( k_{4}, T \right )=1 & \forall & k_{4}\epsilon [2,T]
	\end{matrix}
	\end{equation}
\end{center}

For example, if the key values for start position are $k_{1}$=2 and $k_{2}$=2, respectively and the key values for row and column jump are $k_{3}$=3, $k_{4}$=5, then the co-prime based mapping is depicted in Fig. 3.

\begin{figure}[!htp]
	\begin{center}
		\includegraphics[height=3.8cm,width=0.33\linewidth]{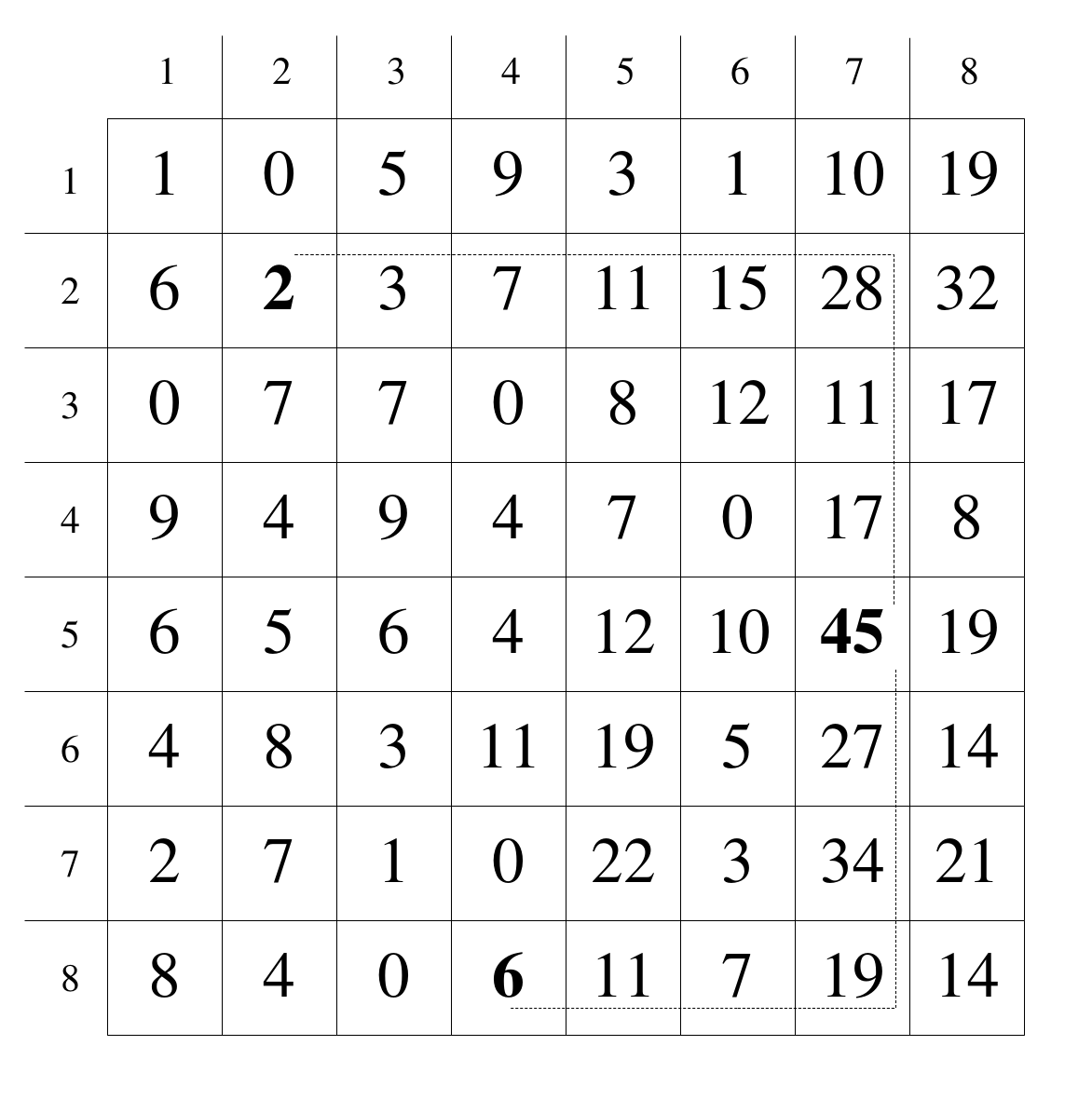}
	\end{center}
	\caption{Example of co-prime based mapping procedure}
\end{figure}

\subsection{Matching}
Fingerprint matching involves the comparison between an enrolled fingerprint template (say $CT$) and a query fingerprint template (say $QT$) to output a match score. In our method, two-step matching is performed to compute match score: Local matching and global matching. 

\subsubsection{Local matching:}
In local matching, ridge feature set corresponding to a minutiae point from $QT$ is compared with ridge feature set for a minutiae point of $CT$ to return local match score. Mapped ridge feature set in the $QT$ and $CT$ are accessed using user-specific keys $k_{1}$, $k_{2}$, $k_{3}$ and $k_{4}$. We compute the Euclidean distance between the mapped non-zero entries of the query and enrolled template. Next, we compute the mean of the minimum distances corresponding to each non-zero entries of the two ridge features sets of $CT$ and $QT$ as described in Eq. (6). 

\begin{equation}
e\_dist = \sqrt{\left ( QT_{N}[i][1]-CT_{M}[j][1] \right )^{2}+\left ( QT_{N}[i][2]-CT_{M}[j][2] \right )^{2}}
\end{equation}

\subsubsection{Global matching:}
In global matching, we compute the number of matched minutiae points between $QT$ and $CT$ utilizing the local match scores by comparing each ridge-feature set from $QT$ with each ridge-feature set from $CT$. Next, overall matching score is evaluated by the number of matched minutiae points divided by the number of minutiae points in $QT$ as described in Eq. (7). 

\begin{equation}
overall\_match\_score=\frac{match\_minutiae\_count}{N}
\end{equation}

\section{Experimental Results and Analysis}
In our experiment, we use four datasets DB1, DB2, DB3 and DB4 of FVC2002 and FVC2004 databases \cite{fvc} since the most of the existing approaches utilized these datasets. Each datasets DB1, DB2, DB3 and DB4 of FVC2002 and FVC2004 contain a total of 800 images of 100 subjects with eight samples each. The performance of the method is evaluated with four parameters: False Acceptance Rate (FAR), False Rejection Rate (FRR), Equal Error Rate (EER) which is defined as the error rate when the FRR and FAR holds equality, and GAR is computed as 1-FRR.

\subsection{Validation of parameter: Number of sectors ($s$)}
The proposed method divides the input fingerprint image into the $s$ number of sectors with equal angular width after preprocessing. To validate the parameter $s$, we have performed a number of experiments considering different values of $s$. We have computed the EER with angular width of $15\degree$, $30\degree$, \ldots and $90\degree$ corresponding to $s$ = 24,12,\ldots and 4, respectively. The performance for the different values of $s$ is reported in Table 1. From the reported results in Table 1, we observe that $s$ =8 corresponds to the best performance on each of the datasets of FVC2002. Also, it has also been observed that EER increases due to more number of sectors without minutiae points for high values of $s$. Therefore, we have considered $s$ = 8 for all other experiments for each dataset of FVC2002 and FVC2004 databases.

\begin{table}[b]
	\caption{EER obtained for databases FVC 2002 DB1, DB2, DB3 and DB4 in same key scenario}
	\centering
	\resizebox{\textwidth}{!}{%
	\begin{tabular}{|c|c|c|c|c|}
		\hline
		\multirow{2}{*}{\begin{tabular}[c]{@{}c@{}}Number of sectors (s)\end{tabular}} & \multicolumn{4}{c|}{EER (in \%)} \\ \cline{2-5} 
		& \begin{tabular}[c]{@{}c@{}}FVC2002 DB1\end{tabular} & \begin{tabular}[c]{@{}c@{}}FVC2002 DB2\end{tabular} & \begin{tabular}[c]{@{}c@{}}FVC2002 DB3\end{tabular} & \begin{tabular}[c]{@{}c@{}}FVC2002 DB4\end{tabular} \\ \hline
		4 & 3.93 & 3.79 & 5.86 & 6.83 \\ \hline
		\textbf{8} & \textbf{1.82} & \textbf{1.39} & \textbf{4.02} & \textbf{5.77} \\ \hline
		16 & 5.04 & 4.93 & 8.83 & 12.7 \\ \hline
		32 & 9.63 & 5.19 & 11.24 & 19.3 \\ \hline
	\end{tabular}}
\end{table}

\subsection{Performance}
We have utilized FVC protocol to evaluate the performance of our method. In this protocol, each subject is compared against the first sample of the remaining subjects to calculate impostor scores. Further, the genuine score is computed by comparing each sample against the remaining samples of the same subject. Hence,  it requires 4950 and 2800 impostor and genuine comparisons for all four datasets of FVC2002 and FVC2004 databases respectively if all samples are enrolled. Further, we have conducted the experiments under two scenarios: Same key scenario and different key scenario.
\subsubsection{Same key scenario:}
This scenario represents the performance under the assumption that all the users hold the same key. In this situation, an adversary utilizes the key being a genuine user to gain unauthorized access to the system. To evaluate the performance, we utilize same keys for all users (i.e. $k_{1}$, $k_{2}$, $k_{3}$ and $k_{4}$) for enrollment. Next, we apply the proposed method to DB1, DB2, DB3 and the DB4 dataset of database FVC2002 and FVC2004. Figure 4 and Figure 5 represent the ROC curves for each dataset of FVC2002 and FVC2004 databases for the optimal value of parameter $s$ (i.e., $s$=8), respectively.
\begin{figure}[!htbp]
	\centering
	\includegraphics[height=7.5cm,width=15cm]{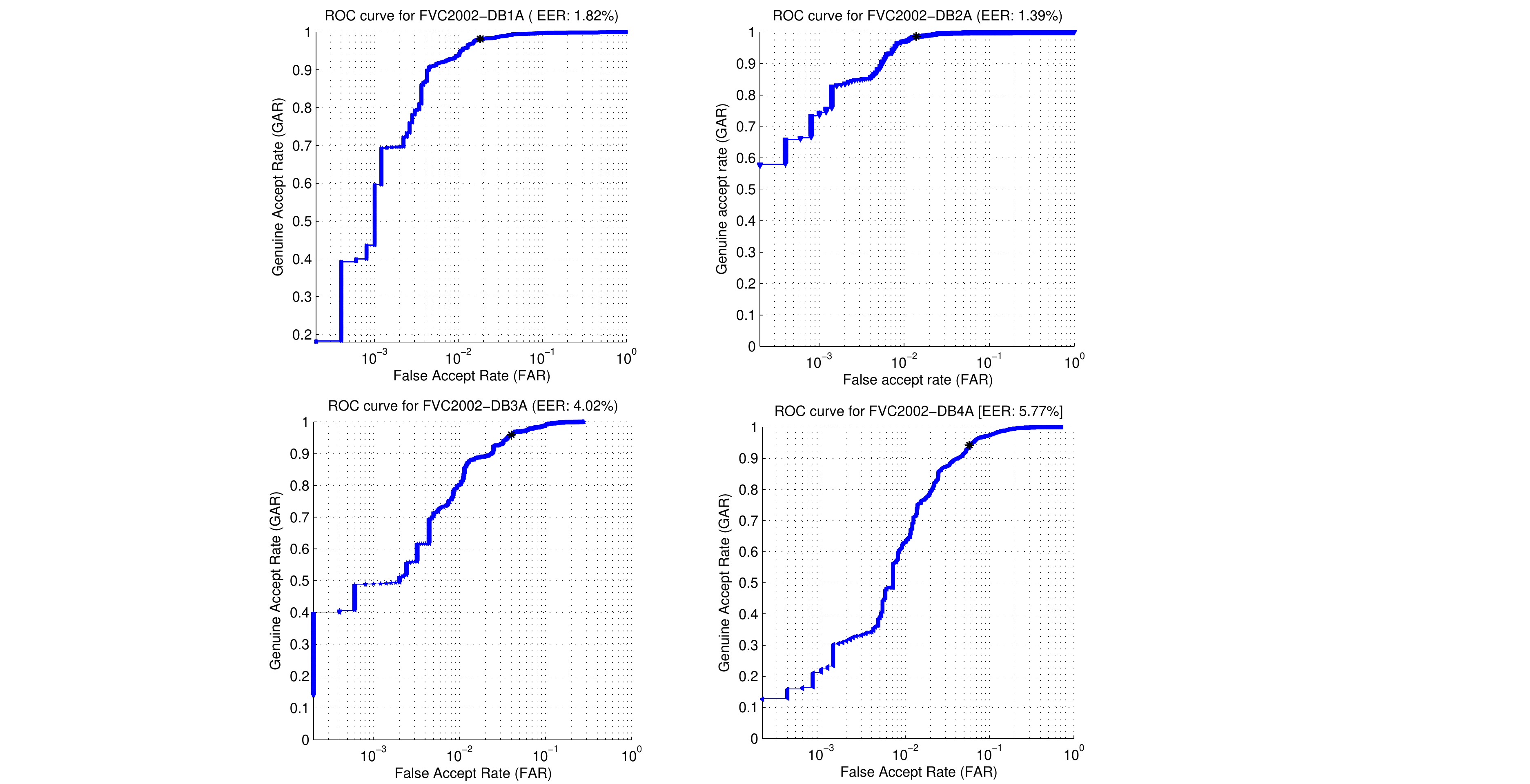}
	\caption{ROC curves for FVC 2002 DB1, DB2, DB3 and DB4 under same key scenario}
\end{figure}\\
\newline
\noindent \textit{FVC2002}: For the database FVC2002, we obtain an EER of 1.82, 1.39, 4.02, and 5.77 for DB1, DB2, DB3, and DB4, respectively under FVC protocol. Out of all four datasets, the proposed scheme attains superior performance onto DB1 and DB2 as these datasets comprise plenty of good quality images in comparison to DB3 and DB4 datasets. Also, DB3 and DB4 datasets include poor quality images containing less number of minutiae points per image as compared to dataset DB1 and DB2. Consequently, we obtain high values of EER for the two datasets, i.e. DB3 and DB4. 

\noindent \textit{FVC2004:} We also apply the proposed method onto the FVC2004 database. As a result, we attain an EER of 8.70, 7.95, 5.23, and 4.87 for DB1, DB2, DB3, and DB4 datasets, respectively. We observe that the method achieves better on DB4, out of all FVC2004 datasets due to relatively better quality images. However, the performance gets degraded in case of DB2 since the first two images of the DB2 dataset are densely distorted. Furthermore, the small overlap area for the images of stored and query templates is another reason for getting less accuracy on FVC2004 DB2. We obtain high values of EER for all four datasets of FVC2004 in comparison to all datasets of the FVC2002 database since deliberate deformation are requested to each user at the time of acquisition \cite{fvc}.

\begin{figure}[!htbp]
	\centering
	\includegraphics[height=7.5cm,width=15cm]{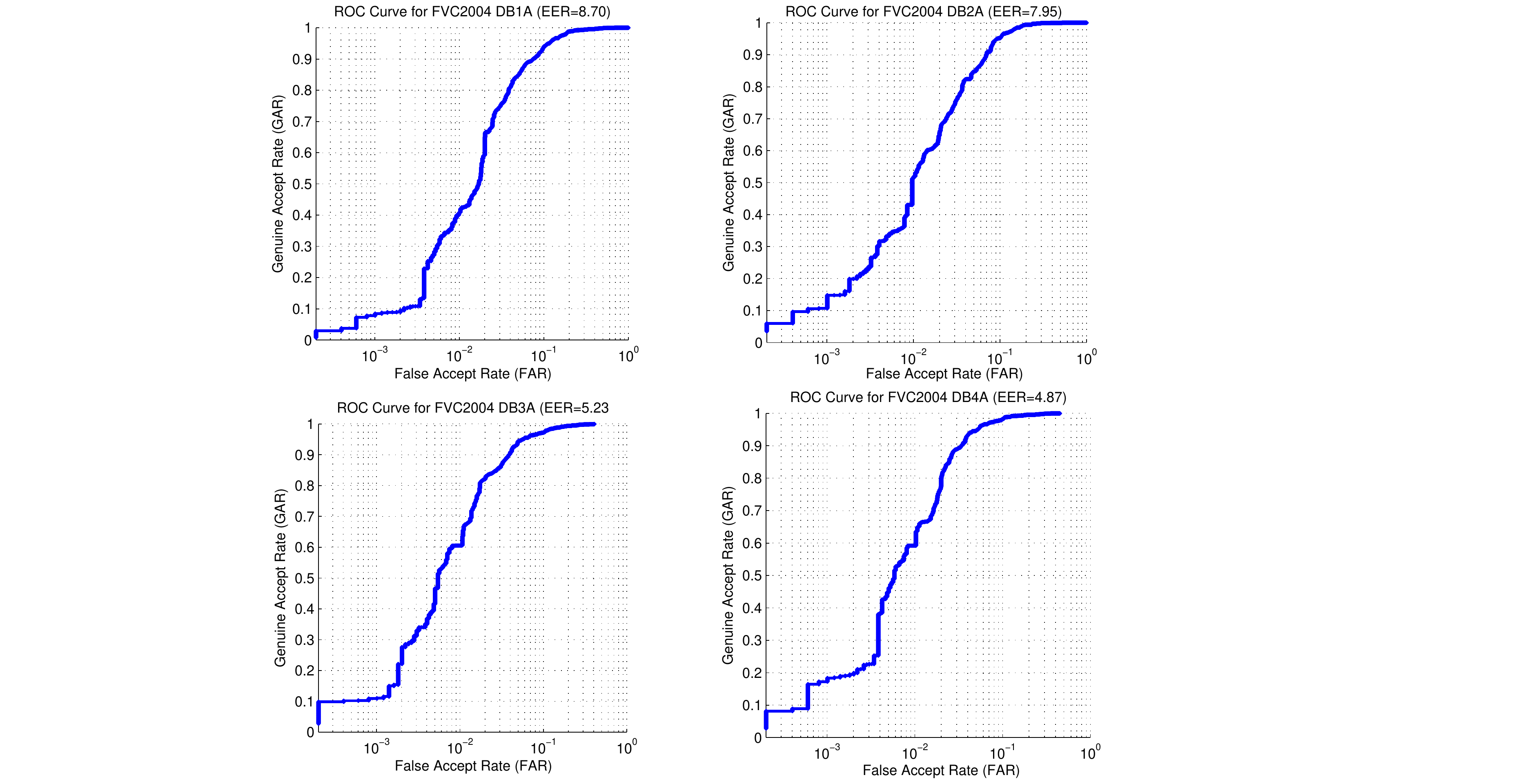}
	\caption{ROC curves for FVC 2004 DB1, DB2, DB3 and DB4 under same key scenario}
\end{figure}

\subsubsection{Different key scenario:}
We also evaluate our method in the scenario where different keys (i.e. $k_{1}$, $k_{2}$, $k_{3}$ and $k_{4}$) are utilized to enroll different users. We achieve an EER of 0 for DB1, DB2 and DB4 datasets and an EER of 0.09 for the dataset DB3 of FVC2002. For FVC2004 database, we obtain an EER of 3.08, 2.25, 1.82 and 1.13 for DB1, DB2, DB3 and DB4 datasets, respectively which is reasonable considering relative poor quality images. Hence, it is clear that our approach also outperforms in the different key scenario.

\subsubsection{Time complexity:}
To investigate the time complexity of our method, we calculate the template generation time and matching time on a machine with specification as, Intel CPU i5, 2400 (3.10 GHz) and 4 GB RAM. The time readings are reported in Table 2. Note that MATLAB 2016b is used without any code optimization. It has been observed from the tabulated results that protected template generation takes almost equal time with the matching time since the matcher has to retrieve the desired mapped positions in different templates for comparison. Overall, the average time elapsed in template generation and matching meets the expectations for an efficient real-time realization. 

\begin{table}[!htbp]
	\centering
	\caption{Average processed time for proposed method under protected template generation and matching}
	\label{my-label}
	\begin{tabular}{|c|c|c|c|c|c|c|c|c|}
		\hline
		\multirow{2}{*}{\begin{tabular}[c]{@{}c@{}}Proposed \\ method\end{tabular}} & \multicolumn{4}{c|}{\begin{tabular}[c]{@{}c@{}}Average time (in sec.)\\ (FVC2002)\end{tabular}} & \multicolumn{4}{c|}{\begin{tabular}[c]{@{}c@{}}Average time (in sec.)\\ (FVC2004)\end{tabular}} \\ \cline{2-9} 
		& DB1 & DB2 & DB3 & DB4 & DB1 & DB2 & DB3 & DB4 \\ \hline
		\begin{tabular}[c]{@{}c@{}}Template\\ generation\end{tabular} & .0005 & 0.0007 & 0.0005 & 0.0006 & 0.0006 & 0.0007 & 0.0005 & 0.0006 \\ \hline
		Matching & 0.0002 & 0.0003 & 0.0002 & 0.0002 & 0.0003 & 0.0002 & 0.0002 & 0.0003 \\ \hline
	\end{tabular}
\end{table}

\subsection{Comparison with existing approaches}
The approaches proposed in \cite{das,shell,ditom,blind,hadamard,pmcc,2pmcc,hamem,jintee,abe,sandhya} utilized FVC2002 database to measure the performance for their method. Furthermore, few methods in current literature such as \cite {lee,hamem,jintee,abe,sandhya} also evaluated their method onto FVC2004 databases. Hence, we have performed the experiments onto FVC2002 and FVC2004 databases and compared the performance of our method with the methods described in \cite{das,shell,ditom,blind,hadamard,pmcc,2pmcc,hamem,jintee,abe,sandhya,lee} to test the efficacy of the proposed method. In Table 3, the proposed method has been compared with the existing in terms of EER. In case of the FVC2002 database, we observe that our method outperforms in comparison to the approaches proposed in \cite{shell,das,ditom,blind,2pmcc,hamem,jintee,abe,sandhya}. Although the performance of our method is slightly lower in case of DB1 for Wang et al. \cite{hadamard}, DB2 for Ferrara et al. \cite{pmcc} and DB4 for Abe et al. \cite{abe} even comparable with existing approaches. For FVC2004 databases, we find that EER obtained by our method is superior to the existing literature as described in Table 3. Hence, it is obvious from Table 3 that our approach outperforms over the current state-of-the-art.
\begin{landscape}
\begin{table}
	\centering
	\caption{EER obtained for databases FVC 2002 DB1, DB2, DB3 and DB4 in same key scenario}
	 \begin{tabular}{|p{0.3\linewidth}|l|l|l|l|l|l|l|l|}
	 	\hline
	 	\multirow{3}{*}{Methods} & \multicolumn{8}{c|}{EER (in \%)} \\ \cline{2-9} 
	 	& \begin{tabular}[c]{@{}l@{}}FVC2002\\ DB1\end{tabular} & \begin{tabular}[c]{@{}l@{}}FVC2002\\ DB2\end{tabular} & \begin{tabular}[c]{@{}l@{}}FVC2002\\ DB3\end{tabular} & \begin{tabular}[c]{@{}l@{}}FVC2002\\ DB4\end{tabular} & \begin{tabular}[c]{@{}l@{}}FVC2004\\ DB1\end{tabular} & \begin{tabular}[c]{@{}l@{}}FVC2004\\ DB2\end{tabular} & \begin{tabular}[c]{@{}l@{}}FVC2004\\ DB3\end{tabular} & \begin{tabular}[c]{@{}l@{}}FVC2004\\ DB4\end{tabular} \\ \hline
	 	\begin{tabular}[c]{@{}l@{}}Das et al.\\ \cite{das}\end{tabular} & 2.27 & 3.79 & - & - &  &  &  &  \\ \hline
	 	\begin{tabular}[c]{@{}l@{}}Moujahdhi et al.\\ \cite{shell}\end{tabular} & 4.28 & 1.45 & - & - &  &  &  &  \\ \hline
	 	\begin{tabular}[c]{@{}l@{}}Wang et al.\\ \cite{ditom}\end{tabular} & 3.5 & 5 & 7.5 & - &  &  &  &  \\ \hline
	 	\begin{tabular}[c]{@{}l@{}}Lee et al.\\ \cite{lee}\end{tabular} & - & - & - & - & 10.3 & 9.5 & 6.8 & - \\ \hline
	 	\begin{tabular}[c]{@{}l@{}}Wang et al.\\ \cite{blind}\end{tabular} & 3 & 2 & 7 & - &  &  &  &  \\ \hline
	 	\begin{tabular}[c]{@{}l@{}}Wang et al.\\ \cite{hadamard}\end{tabular} & 1 & 2 & 5.2 & - &  &  &  &  \\ \hline
	 	\begin{tabular}[c]{@{}l@{}}Ferrara et al.\\ \cite{pmcc}\end{tabular} & 1.88 & 0.99 & 5.24 & 4.84 &  &  &  &  \\ \hline
	 	\begin{tabular}[c]{@{}l@{}}Ferrara et al.\\ \cite{2pmcc}\end{tabular} & 3.3 & 1.8 & 7.8 & 6.6 &  &  &  &  \\ \hline
	 	\begin{tabular}[c]{@{}l@{}}Jin et al.\\ \cite{hamem}\end{tabular} & 4.36 & 1.77 & - & - & 17.89 & 17.11 & - & - \\ \hline
	 	\begin{tabular}[c]{@{}l@{}}Jin et al.\\ \cite{jintee}\end{tabular} & 5.19 & 5.65 & - & - & 15.76 & 11.64 & - & - \\ \hline
	 	\begin{tabular}[c]{@{}l@{}}Abe et al.\\ \cite{abe}\end{tabular} & 2.3 & 1.8 & 6.6 & 5.1 & 13.4 & 8.1 & 9.7 & 6.3 \\ \hline
	 	\begin{tabular}[c]{@{}l@{}}Sandhya et al.\\ \cite{sandhya}\end{tabular} & 3.96 & 2.98 & 6.89 & - & 12.17 & 13.29 & 17.73 & - \\ \hline
	 	\textbf{Proposed method} & \textbf{1.82} & \textbf{1.39} & \textbf{4.02} & \textbf{5.77} & \textbf{8.70} & \textbf{7.95} & \textbf{5.23} & \textbf{4.87} \\ \hline
	 \end{tabular}%
    \begin{tablenotes}\small
		\item $`-'$ indicates that the author(s) have not reported the results or results are reported for the partial dataset, in their work.
	\end{tablenotes}
\end{table}
\end{landscape}

\subsection{Baseline comparison}
We perform two sets of experiments for baseline comparison. In the first experiment, we evaluate the EER using the original (untransformed) template involving ridge features. In the second experiment, we evaluate the performance after applying the proposed transformation using the keys ($k_{1}$, $k_{2}$, $k_{3}$ and $k_{4}$). In case of FVC2002, the performance is degraded by 0.19\%, 0.41\%, 0.05\%, and 0.39\% for DB1, DB2, DB3, and DB4 dataset of FVC2002 database, respectively. For FVC2004 database, the performance is degraded by 0.03\%, 0.03\%, 0.05\% and 0.06\% for DB1, DB2, DB3, and DB4 datasets, respectively with respect to original (untransformed) fingerprint template. From the reported results in Table 4, it is clear that the proposed cancelable transformation exhibits very low performance degradation.

\begin{table}[!htbp]
	\centering
	\caption{Baseline comparison for FVC2002 database}
			\begin{tabular}{|c|c|l|l|l|c|l|l|l|c|l|l|l|}
			\hline
			\multirow{2}{*}{Database} & \multicolumn{8}{c|}{EER (in \%)} & \multicolumn{4}{c|}{\multirow{2}{*}{\begin{tabular}[c]{@{}c@{}}Performance \\ degradation (in \%)\end{tabular}}} \\ \cline{2-9}
			& \multicolumn{4}{c|}{\begin{tabular}[c]{@{}c@{}}Without cancelable\\ transformation\end{tabular}} & \multicolumn{4}{c|}{\begin{tabular}[c]{@{}c@{}}With cancelable\\ transformation\end{tabular}} & \multicolumn{4}{c|}{} \\ \hline
			FVC2002DB1 & \multicolumn{4}{c|}{1.47} & \multicolumn{4}{c|}{1.82} & \multicolumn{4}{c|}{0.19} \\ \hline
			FVC2002DB2 & \multicolumn{4}{c|}{0.89} & \multicolumn{4}{c|}{1.39} & \multicolumn{4}{c|}{0.41} \\ \hline
			FVC2002DB3 & \multicolumn{4}{c|}{3.81} & \multicolumn{4}{c|}{4.02} & \multicolumn{4}{c|}{0.05} \\ \hline
			FVC2002DB4 & \multicolumn{4}{c|}{3.49} & \multicolumn{4}{c|}{5.77} & \multicolumn{4}{c|}{0.39} \\ \hline
			FVC2004DB1 & \multicolumn{4}{c|}{8.43} & \multicolumn{4}{c|}{8.70} & \multicolumn{4}{c|}{0.03} \\ \hline
			FVC2004DB2 & \multicolumn{4}{c|}{7.69} & \multicolumn{4}{c|}{7.95} & \multicolumn{4}{c|}{0.03} \\ \hline
			FVC2004DB3 & \multicolumn{4}{c|}{4.94} & \multicolumn{4}{c|}{5.23} & \multicolumn{4}{c|}{0.05} \\ \hline
			FVC2004DB4 & \multicolumn{4}{c|}{4.58} & \multicolumn{4}{c|}{4.87} & \multicolumn{4}{c|}{0.06} \\ \hline
	\end{tabular}
\end{table}

\section{Security analysis}
A template protection mechanism should fulfill the requirements of irreversibility, revocability, and diversity as depicted in Section 1. In the following subsections, we analyze our method concerning these criteria. 

\subsection{Irreversibility analysis}
The requirement of irreversibility or non-invertibility states that it should be infeasible to obtain the original template from the protected template. To validate the claim of non-invertibility, we imagine that an adversary unveils the stored protected template $CanTemp$. In this situation, the adversary cannot be able to access the original template ($F$) as he does not have any clue about the four different keys used for mapping. For instance, if an input fingerprint containing 50 minutiae points is divided into eight sectors, then the original template ($F$) and protected template $CanTemp$ would contain 800 cells and 640000 cells, respectively. Hence, It is not feasible to calculate initial positions ($k_{1}$, $k_{2}$) and next positions ($k_{3}$, $k_{4}$) to extract the entries corresponding to original template as an adversary would require $640000\times 640000$ =409 billion attempts.

Further, assume that the attacker reveals the keys ($k_{1}$, $k_{2}$, $k_{3}$ and $k_{4}$) utilized for mapping. In this situation, it would be extremely hard for an attacker to construct original template since all four keys $k_{1}$, $k_{2}$, $k_{3}$ and $k_{4}$ comprise of random coprime entries. From the random coprime entries, it is almost impossible to reveal any information about the original template.

\subsection{Revocability analysis}
The criteria of revocability refer to issue a new template in case stored template gets compromised. To validate the claim of revocability, we have generated 100 different protected templates by altering the parameters for the same fingerprint. Next, genuine, imposter and Pseudo-imposter distribution are computed for the two different datasets of each database i.e. DB2 dataset of FVC2002 and the DB1 dataset of FVC2004. In this experiment, we obtain 0\% average FAR. Also, the mean and standard deviation ($\mu;\sigma$) of genuine, imposter and pseudo-imposter are (0.576;0.02), (0.202;0.031), and (0.209;0.0373), respectively for DB2 of FVC2002. In a similar manner, we achieve (0.528;0.017), (0.196;0.029), (0.189;0.031) for genuine, imposter and pseudo-imposter distributions for DB1 dataset of FVC2004, respectively. From the evaluated distribution, it has been observed that there is a strong overlap between the pseudo-impostor and impostor distributions as shown in Fig. 6. This means that the derived templates from the same subject with different keys are different enough with each other. Hence, it is confirmed that the compromised template differs from the transformed template yet belong to the same fingerprint.

\begin{figure}[!htp]
	\begin{center}
		\includegraphics[height=6cm,width=\textwidth]{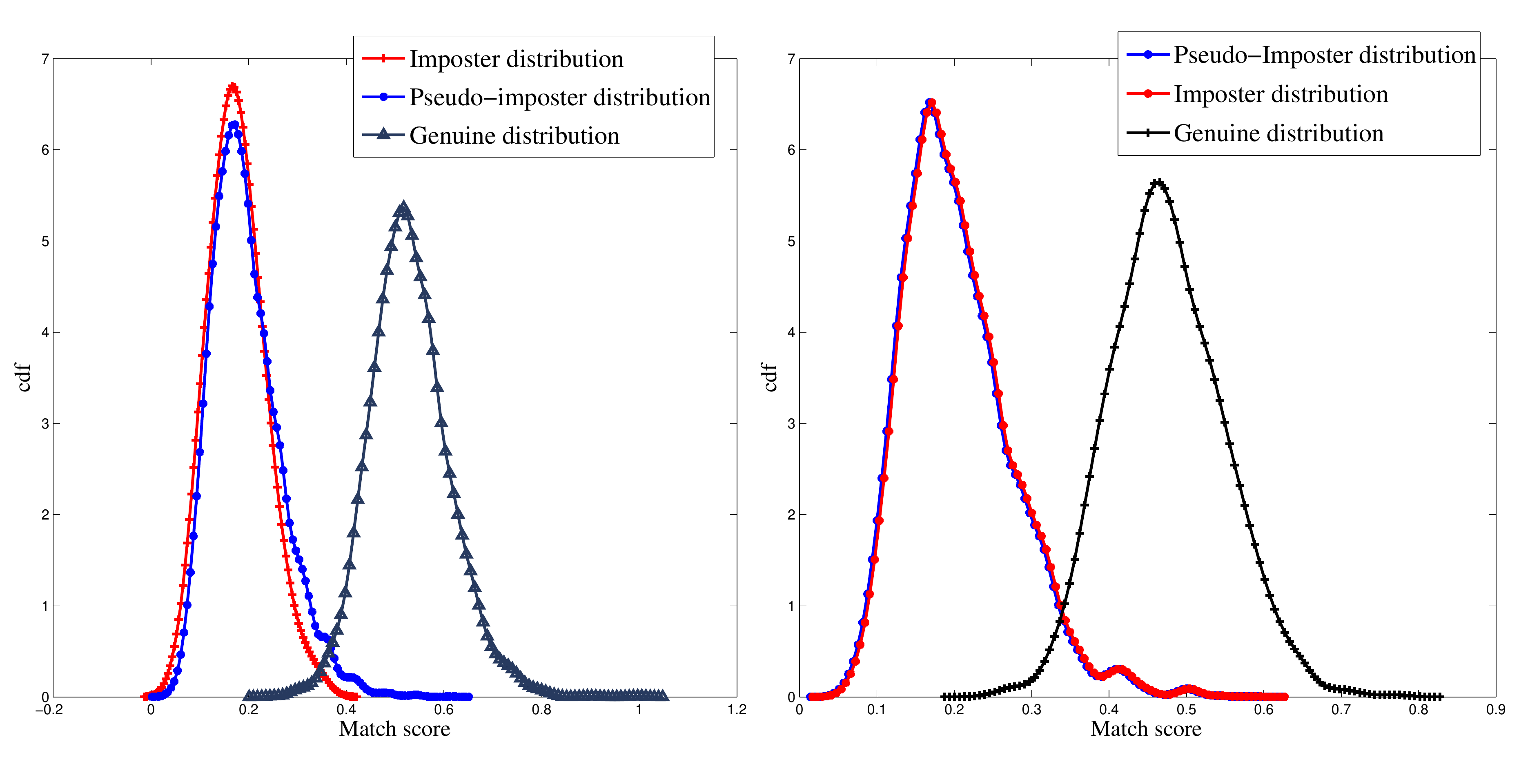}
	\end{center}
	\caption{Genuine, imposter and pseudo-imposter distribution for FVC2002 DB2 (Left) and FVC2004 DB1 (Right)}
\end{figure}

\subsection{Unlinkability analysis}

This characteristic implies that there must be significant distinctiveness between the two protected templates derived from same or different subjects. To validate this requirement, we evaluate pseudo-genuine scores. Pseudo-genuine score is evaluated by comparing two different templates of same subject using different values of $k_{1}$, $k_{2}$, $k_{3}$ and $k_{4}$. 

In this context, if the pseudo-genuine and pseudo-imposter distribution get overlapped, it implies that the protected templates derived from the same subject are adequately dissimilar and vice versa. The computational hardness in differentiating the protected templates aids to the unlinkability characteristics. Figure 7 shows the pseudo-genuine and pseudo-imposter distribution for the DB2 dataset of FVC2002 and the DB1 dataset of FVC2004 where the pseudo-imposter and pseudo-genuine distribution are substantially overlapped. This confirms the unlinkability for derived protected templates.

\begin{figure}[!htp]
	\begin{center}
		\includegraphics[height=7cm,width=\textwidth]{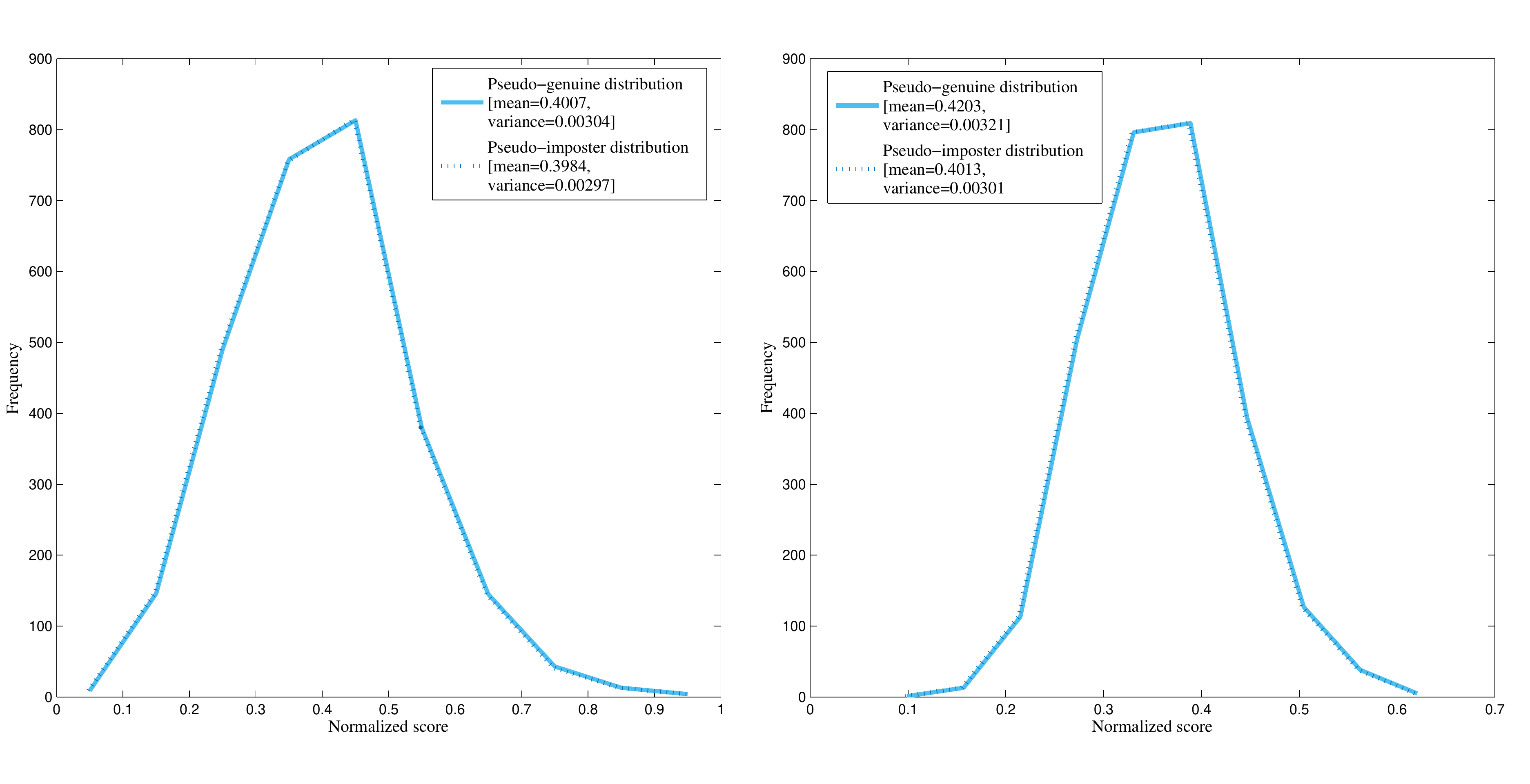}
	\end{center}
	\caption{Pseudo-genuine, pseudo-imposter for DB2 of FVC2002 (left) and DB1 of FVC2004 (right)}
\end{figure}

\subsection{Diversity analysis}
The template generation scheme should be able to derive multiple templates different enough with each other so that it does not cross-match over different applications. Many different templates can be derived by changing the values of key ($k_{1}$, $k_{2}$, $k_{3}$, $k_{4}$) and seed ($\rho$). Further, the number of sectors (\textit{s}) can also be varied to generate multiple templates.

\subsection{Other attacks}
We also analyze our method against different types of attacks namely brute-force, pre-image and annealing attacks to validate the robustness of the proposed work:

\noindent \textit{Brute-Force Attack:}
Brute-force attack determines the total number of attempts made by an imposter to retrieve the original template. This attack is also known as masquerade attack \cite{masq} in literature. In our method, we have achieved the best accuracy at \textit{s}=8. For best performance configuration, an adversary would require 409 billion brute-force attempts to guess the positions of the original ridge features corresponding to an input fingerprint containing 50 minutiae points as described in Section 4.1.

\noindent \textit{Pre-image attack:}
In pre-image attack, the adversary may use multiple instances of the protected template to derive a pre-image of the original template. Privacy invasion is attempted by utilizing feature order in the different protected template to create a fake template. In literature, Biohashing based methods \cite{biohash,hash,hash2} are vulnerable to this attack since they derive a binary string which can be easily exploited to unveil original minutiae information. On the contrary, our method is robust enough to sustain this attack since the coprime mapping is utilized to hide the original ridge features across many possible different non-overlapped coprime positions. Also, our method does not depend on the order of feature components while generating the original as well as the protected template. Further, any value could not be investigated from the two projected feature vectors in any position due to the different sized enrolled and query template. Hence, pre-image attack could not be utilized to derive the original template in our method.    

\noindent \textit{Annealing attack:} 
In this attack \cite{anneal}, the protected template is divided into multiple regions, and some regions of a sample template are paired with some regions of the reference template to evaluate similarity score. If the similarity score exceeds the threshold, the vicinity corresponding to sample's region is included in the gummy template. This step is repeated until it outputs a gummy template including all the matched vicinities. Our approach is robust against this type of attack due to the following reasons:
\begin{enumerate}
	\item Our approach evaluates the nearest-neighbor minutia for each minutiae point causing the different radii to the different minutiae points. Hence, it is very hard to map the gummy template with the original template which is derived from the multiple regions with the variable radius.
	\item Ridge-based features are utilized for the neighboring minutiae points in each sector instead of relative distances or the directional difference between minutiae pairs. Here, the measured ridge features are invariant to the inter-ridge distances and locations of minutiae points. 
\end{enumerate}

\section{Conclusion}
Pre-alignment and the privacy-invasion are two prime factor in fingerprint-based authentication. To address these issues, we have proposed a novel cancelable fingerprint template generation method which maps original ridge features to a coprime position in a non-overlapping manner. The approach does not depend on detection of singularities (core/delta). Due to the simplicity in implementation, the method is suitable for real-world applications such as mobiles and smart- cards. The experimental evaluations performed over two publicly available databases FVC2002, and FVC2004 databases show a significant performance improvement in comparison to the several existing works in fingerprint template protection. Further, the necessary criteria of non-invertibility, diversity, and revocability are either theoretically proved or verified by experiments. The proposed method is tested against different attacks and observed that the transformation ensures the optimal security and preserves the recognition accuracy. In the future, we would try to invent a method for the evaluation of ridge feature for low or poor quality fingerprint and partial fingerprint images. 

\bibliographystyle{spphys}       
\bibliography{egbib}   


\end{document}